\documentclass[letterpaper, 10 pt, conference]{ieeeconf}  

\IEEEoverridecommandlockouts                              

\usepackage{amsmath,amsfonts,amssymb}
\usepackage{mathtools}
\usepackage{commath}

\usepackage{array}
\usepackage{bm}
\usepackage{booktabs}
\usepackage{multirow}

\usepackage{subcaption}
\usepackage{wrapfig}

\usepackage{multicol}
\usepackage[bookmarks=true]{hyperref}
\usepackage{tabularx} 
\usepackage{caption}
\usepackage{graphicx}

\usepackage{authblk}
\newcommand{\pcx}[1]{X^{(#1)}}
\newcommand{\wxy}[2]{W_{#1 \rightarrow #2}}

\title{\LARGE \bf
 One-Shot Cross-Geometry Skill Transfer through Part Decomposition 
}
\author[1]{Skye Thompson}
\author[2]{Ondrej Biza}
\author[1]{George Konidaris}

\affil[1]{Brown University}
\affil[2]{Northeastern University}

\date{}                     
\begin{document}

\maketitle
\thispagestyle{empty}
\pagestyle{empty}

\begin{abstract}

Given a demonstration, a robot should be able to generalize a skill to any object it encounters—but existing approaches to skill transfer often fail to adapt to objects with unfamiliar shapes. Motivated by examples of improved transfer from compositional modeling, we propose a method for improving transfer by decomposing objects into their constituent semantic parts. We leverage data-efficient generative shape models to accurately transfer interaction points from the parts of a demonstration object to a novel object. We autonomously construct an objective to optimize the alignment of those points on skill-relevant object parts. Our method generalizes to a wider range of object geometries than existing work, and achieves successful one-shot transfer for a range of skills and objects from a single demonstration, in both simulated and real environments 

\end{abstract}

\section{INTRODUCTION}
Given a demonstration of a manipulation skill on a given object, a robot should be able to generalize that skill to novel objects---even objects that that vary substantially in shape from those it has seen before. But existing approaches fail to generalize to object geometries substantially outside their training context, especially when few demonstrations or object examples are available. Few-shot imitation learning methods that attempt to leverage geometric priors \cite{simeonov21ndf} \cite{biza23oneshot} have shown some success in enabling transfer, but defining a prior that captures the full range of variations in object shape is nontrivial, resulting in misprediction and transfer failure if that prior is misaligned with an object instance. 

Successful skill adaptation requires the ability to both identify the features of a novel object that are salient for a given skill, and to successfully adapt that skill policy to the novel object using those features. Object geometric features are complex, high dimensional, and highly variable even within a category. Those features can also have a complex relationships to successful skill execution---for example, the trajectory needed to pour from a teapot with a long spout and short handle may differ significantly from a teapot with a short spout and a long handle. Existing skill learning approaches fail to capture this complexity. How can we better represent and condition on objects' geometric features to enable skill transfer to a wider variety of objects?

\begin{figure}[]
\includegraphics[width=.95\columnwidth]{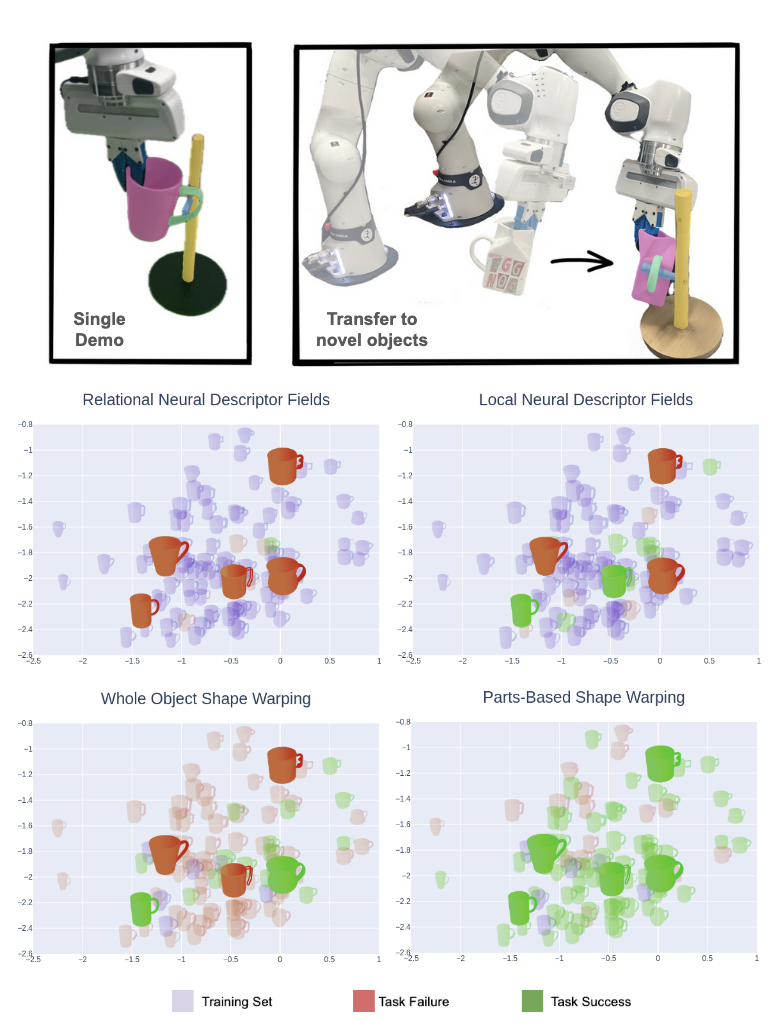}
\captionof{figure}{Using a decomposed object representation, parts-based shape warping generalizes a manipulation skill to the widest range of object geometries from the fewest training examples. We show results on a simulated mug-on-rack task, with a subsample of tested mugs enlarged for comparison. }
\label{fig:opener}
\vspace{-3ex}
\end{figure}

Motivated by examples of improved transfer from compositional modeling \cite{liu2024composablepartbasedmanipulation} \cite{chun2023local}, we explore a method for improving transfer performance using decomposition---representing objects as compositions of independent representations, rather than a single, monolithic representation. Existing work \cite{DBLP:journals/corr/abs-1812-02713} \cite{qian2024taskorientedhierarchicalobjectdecomposition} presents us with a natural decomposition for many objects and skills: part decomposition, representing part shape and relationships independently. Rather than model each object monolithically, we instead model each object as reconstructed from models of its segmented constituent parts---for example, modeling a handle, body, lid, and spout, rather than a monolithic teapot. This decomposition improves transfer in two ways; first, it disentangles the learned feature representation of different parts from the relationship between those parts, which enables more accurate shape reconstruction and feature identification. And second, it enables skill generalization to a wider range of objects by modeling the relationships between object parts independently, adapting effectively when those part relationships change. 

To evaluate this approach, we propose a method for parts-based feature transfer and skill conditioning that adapts a demonstrated manipulation skill to a range of novel objects of different shapes. We leverage existing semantic segmentation models \cite{kirillov2023segment} to decompose objects into a fixed set of part categories. Given a single demonstration of a skill, our approach identifies relevant features for each part of each object in the demonstration. We transfer the part features to the novel object using generative shape models, extending existing work for transforming object features via shape warping \cite{biza23oneshot}. These transformed features are then used to autonomously compose an optimization objective in the novel scene that adapts the demonstrated skill with no additional training.

We compare performance on two benchmark tasks in simulation---placing a mug on a rack, and a bowl on a mug---with previous work that does not leverage part decomposition, and existing work utilizing neural descriptor fields with and without locally conditioned features  \cite{simeonov21ndf} \cite{chun2023local} (a different approach leveraging a similar parts-focused insight). We compare over a range of object geometries and provided demonstrations. Part decomposition allows us to successfully transfer a skill to a wider variety of objects than existing methods. We also evaluate this approach in real world pouring and placing tasks, using real objects from categories not well represented in existing benchmark datasets like ShapeNet \cite{shapenet}, demonstrating that this policy representation can achieve accurate transfer from limited examples.

\section{Skill Transfer from Demonstration}

We consider the task of manipulating a rigid object $A$ into a specific pose in relation to a rigid object $B$. We define a successful skill execution as achieving an SE(3) transformation $T_{AB}$, in world frame ${W}$, that transforms $A$ from its initial position to a final pose relative to $B$ that achieves the task goal, such that task cost $C(A, B, T_{AB}) = 0$. Given a demonstration of a skill performed on these two objects, we want to transfer that demonstration to two previously unseen test objects from the same categories, $A'$ and $B'$. These may differ in both initial pose and geometry from the demonstration objects, impacting the transformation $T_{A'B'}$ required to achieve task success. 

We assume this transformation is a function of relevant shape features grounded in the object's SE(3) reference frame.Concretely, we say there exists a mapping  $A \rightarrow I^{(A)}$, where $I^{(A)}$ refers to a set of features grounded to the object's geometry. Each feature is grounded by associating a features $i^{(A)} \in I^{(A)}$ with a point $(x^{(A)}_i, y^{(A)}_i, z^{(A)}_i)$ in the object's reference frame. This definition covers both point features, like key points \cite{manuelli19kpam} or contact points \cite{rodriguez18transferringa}, and  semantic or shape features associated with a point or specific SE(3) transformation \cite{simeonov21ndf}, \cite{biza23oneshot}. 

Given a demonstration, we can map these features to a transformation that successfully completes the skill, $\pi(I^{(A)}, I^{(B)}) = T_{AB}$, which can then be achieved by a motion planner or other fixed controller. Our goal is to capture a mapping $\pi$ that transfers to \textit{novel} objects, meaning it also captures $\pi(I^{(A')}, I^{(B')}) = T_{A'B'}$ such that $C(A', B', T_{A'B'}) = 0$.

\subsection{Decomposition and Warping for Skill Transfer} \label{mymath}

Warping is a class of solutions to the problem of skill transfer that assume there is a correspondence between the transformation of features observed in different environments $f(I^{(A)}, I^{(B)}) = (I^{(A')}, I^{(B')})$ and the transformation between policies that succeed in those environments $g(T_{AB}) = T_{A'B'}$. (We will shorten $f(I^{(A)}, I^{(B)})$ as $f(A,B)$ for brevity, assuming that $I^{(A)}$ is fully determined by object geometry $A$). Given $f(A,B)$ and a functional mapping of the feature transformation to the policy transformation, $f \rightarrow g$, we can then predict $T_{A'B'}$ given $(A, B)$, $(A', B')$, and $T_{AB}$---which constitute a demonstration in a known environment, and an observation of relevant features in the novel environment. Following Schulman et al. \cite{Schulman2016}, we pose this problem of transfer learning in manipulation as one of cost invariance. We assume a global task cost function $C(s, T)$, where $s$ is the initial state of the environment, and $T$ is the trajectory the robot takes to complete the skill. Shape warping is a successful strategy for achieving skill transfer when

\begin{equation}
\label{cost_invariance}
C(s, T)  = C(f(s),  g(T)) = 0
\end{equation}
given $f \rightarrow g$. That is, transfer is successful when $f$ and $g$ represent a smooth class of transformations to which skill cost is invariant at least for cost function values close to zero. This is a compelling model for skill transfer when $f$ and $g$, and their mapping, can be effectively modeled with the available data. Early approaches achieved this using nonparametric techniques to infer $f$, and applying a fixed mapping $f \rightarrow g$ (in the case of Schulman et al. \cite{Schulman2016}, $f = g$) to transform the manipulator trajectory. Other works, like \cite{rodriguez18transferringa}, learn the mapping $f \rightarrow g$ from a larger set of skill demonstrations.  However, accurately capturing this mapping from limited examples is often difficult, resulting in failure to generalize to novel object examples. 

Later work attempts to address this by defining $g$ as not a single transform, but as an optimization over transforms given a substitute heuristic---e.g. 
\begin{equation}
g(T) = \operatornamewithlimits{argmin}\limits_{T'}(h(f(A,B), T')),
\end{equation}
where $h$ is some heuristic more easily observed than the true task cost of executing $g(T)$. This heuristic objective can be learned, as in Thompson et. al. \cite{thompson21shapebased}, or prescribed, as in Biza et al. \cite{biza23oneshot} where it is represented by the alignment of nonparametrically discovered keypoints, or Simeonov et al. \cite{simeonov21ndf} where it is the alignment of sampled neural descriptors. These methods enable better transfer from limited data because posing the mapping $f \rightarrow g $ as an optimization problem broadens the scope of transformations that can achieve the cost invariant condition---therefore, the necessary condition for successful transfer becomes 
\begin{equation}
\label{heuristic}
C((A, B), T) = C(f(A, B), \operatornamewithlimits{argmin}\limits_{T'}(h(f(A,B), T'))),
\end{equation}
for small values of $C$---more easily obtained in amenable optimization environments, given that minimizing $h$ correspondingly minimizes $C$.

\section{ Improved Transfer through Part Decomposition }

\subsection{Applying Part Decomposition to Skill Transfer}
In this work, we consider the benefit of part decomposition in this problem setting---when a suitable $h$ is not well defined over whole objects, we instead consider optimization over a composition of heuristics conditioned on the features of individual objects parts $m \in M$ of $A$ and $n \in N$ of B:
\begin{equation}
\label{heuristic_parts}
C((A, B), T)  = C(f(A, B), \operatornamewithlimits{argmin}\limits_{T'}\sum_{m,n}^{M,N}{(h(f(m,n), T')})).
\end{equation}

To apply this in a skill transfer setting, first, we must infer features for decomposed parts. Formally, rather than trying to learn $f = I^{(A)} \rightarrow I^{(A')}$, for a category $A$ with a fixed set of component part categories $M$, we learn a set of $f_m = I^{(m)} \rightarrow I^{(m')} $ for all $m \in M$. Each $f_m$, where $f(A) =  \bigcup\limits_{m}^{M} f_{m}(m)$, is likely to be simpler, and to represent a smoother class of transformations, than monolithic $f$, making for more accurate inference from a learned model. 

\begin{figure}[h]
    \includegraphics[width=.95\columnwidth]{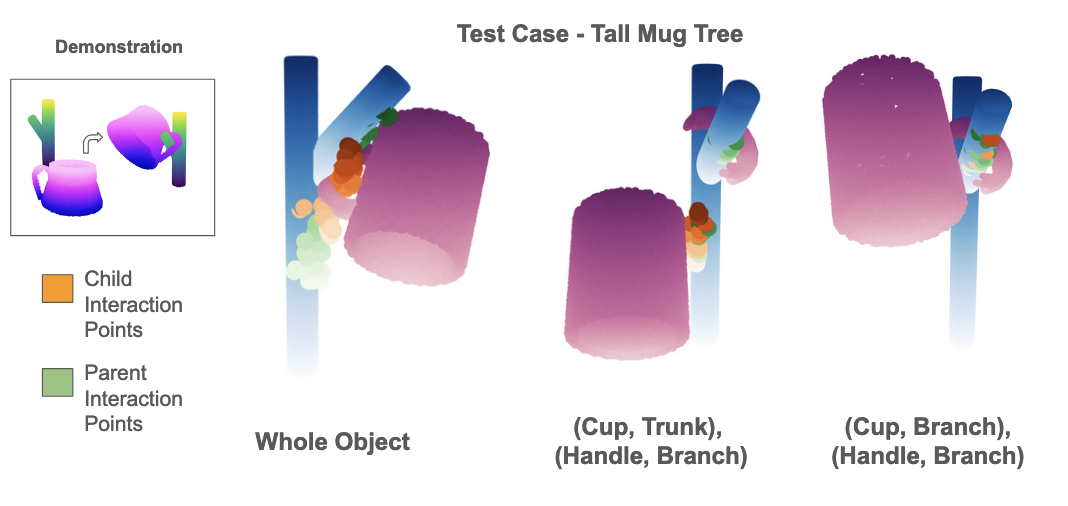}
    \caption{Conditioning a skill trajectory on keypoints transferred to a novel object fails when the part relationship between the demonstration and novel object changes. Part decomposition improves the accuracy of feature transfer, but still fails to adequately adapt the skill. Identifying and recomposing relevant part relationships better transfers features and successfully completes the skill on the novel object. }
    \label{fig:relevant_parts}
    \vspace{-3ex}
\end{figure}

As demonstrated in Figure \ref{fig:relevant_parts}, our second motivation is that part decomposition allows us to condition our trajectory selection on only parts of the object that are determined to be salient to the skill being transferred and amenable as components for our heuristic optimization. We can leverage these properties to achieve successful transfer for a wider range of objects with a wider variety of features.

Given object categories $A$ and $B$ with part categories $M$ and $N$ we make the heuristic substitution from equation \ref{heuristic} to replace equation \ref{cost_invariance} with the following:
$$ C(f(A,B), { \operatornamewithlimits{argmin}\limits_{T'}(\sum_{m^*,n^*}h(f_{m^*,n^*}(A, B), T')}),$$
where $(m^*, n^*) \in M^* \prod N^* $, and are identified  at training time to best maintain this cost invariant property based on their accuracy in reconstructing known examples:
$$M^*, N^* = \operatornamewithlimits{argmin}\limits_{M^* \subset M, N^* \subset N}(\hat{C}(A, B, T_{AB}) - C((A,B), T). $$
$\hat{C}$ is the cost of a set of one or more example objects and trajectories---we consider a single demonstration. This can be combinatorially expensive to evaluate depending on the size of $M$ and $N$. We consider small $M$ and $N$ in this work. Tree pruning or program induction techniques could be useful for making larger search spaces manageable.

\begin{figure*}
    \centering
    \includegraphics[width=.9\textwidth]{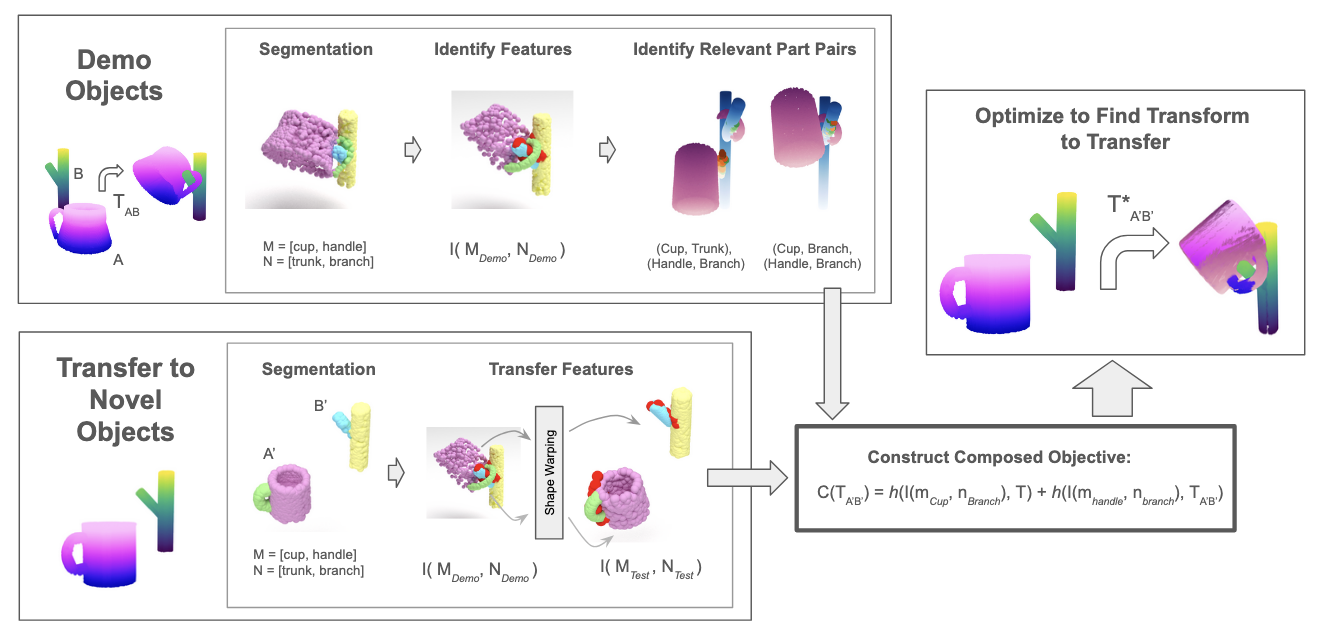}
    \caption{The full pipeline for skill transfer through part decomposition. }
    \label{fig:big_fig}
    \vspace{-3ex}
\end{figure*}

For our evaluation, we focus on transferring skills to objects of different geometry. Given a single demonstration of a manipulation skill on objects $A$ and $B$, our goal is to transfer that skill to two novel objects from the same categories, $A'$ and $B'$. Processing the provided demonstration requires three stages---1) identifying the objects in the demonstration, 2) inferring their geometry, and 3) inferring a set of annotations that capture the demonstration transformation $T_{AB}$. Given a fixed decomposition of category $A$ into parts $M$, and $B$ into $N$, we address how to perform each process on the object's \textit{parts}, to transfer the extracted part relationships to the novel objects. We then address reconstructing the transferred policy that fulfills those relationships, and executing it to successfully complete the skill.

We begin with a demonstration consisting of initial point clouds $X^{(A)}$ and $X^{(B)}$ and final object transform $T_{(AB)}$. We decompose the objects into parts $X^{(M)}$ and $X^{(N)}$, and identify a set of grounded interaction points $I^{(m)}_{n}$  and $I^{(n)}_{m}$ for $(m,n) \in M,N$ that represent the relationship between those object parts in the demonstration. 

\subsection{Identifying and Transferring Part Features}

We assume known categories of objects $A$ and $B$ involved in the demonstrated skill, and known part categories per category, $M$ and $N$. For instance, we assume knowledge of a category \textit{mug} that decomposes into parts \textit{cup} and \textit{handle}. Such decompositions are well-studied in existing work on segmentation \cite{DBLP:journals/corr/abs-1812-02713} \cite{wei2023ovparts} \cite{qian2024taskorientedhierarchicalobjectdecomposition}. Given a mesh, pointcloud, or an RGB-D image of an initial scene in which to execute the manipulation skill on object instances $A'$ and $B'$, we extract the segmented pointclouds associated with each object part, $X^{(m)}$ and $X^{(n)}$ for every $(m, n) \in (M, N)$. In simulation, we use hand-engineered segmentation methods. Our real-world segmentation pipeline is described in Section \ref{segmethods}.

We next need to identify grounded features $I^{(A)}$ and $I^{(B)}$ from which the policy transformation $T_{AB}$ that completes the skill can be determined. For this work, we use interaction warping\cite{biza23oneshot}, selecting aligned pairs of nearby points on each object's pointcloud from which the relative transform between objects can be reconstructed. Rather than consider interaction points between the pair of objects, at training time, we identify interaction points between \textit{all unique pairs of parts} across the two interacting objects.

Given interaction points  $I^{(m)}$ and $I^{(n)}$, and partial views of our novel object parts $X^{(m')}$ and $X^{(n')}$, a model is required to determine transformation $f$ to the equivalent positions of said annotations on the novel parts, $I^{(m')}$ and $I^{(n')}$. We achieve this by using the existing shape-warping method described in Rodriguez et al.\cite{rodriguez18transferringa}, Thompson et al. \cite{thompson21shapebased}, and Biza et al. \cite{biza23oneshot}, which learns a generative model for inferring object geometry and the positions of interaction points on novel objects given a partial point cloud view, by searching in the latent space learned by PCA over 5-10 pose-aligned point cloud segments. This approach is well-described in Biza et al. \cite{biza23oneshot}; implementation details can also be found in Section \ref{iwstuff2}. As above, rather than represent whole objects with a single model, we train a generative model per object part category.

\subsection{Improved Reconstruction with Relational Descriptors}

Decomposed representations offer simplicity and specificity, but at the cost of losing relational information. A monolithic model of a mug is likely to capture the consistent orientation of a mug handle relative to the mug's cup. But without this information, we found our decomposed generative models frequently produced solutions in unfavorable local minima, due to their indifference to axes of symmetry or deformation of the reconstructed shape beyond the observed points. This resulted in inaccurate estimations of $I^{(m')}$ and $I^{(n')}$, particularly for parts like handles with narrow or occluded features. 

We can leverage both the benefits of decomposition and relational information by considering the lost information when doing inference---in this case, the knowledge that our reconstructed shapes come from parts of rigid objects, which have structured relationships to each other that are largely similar within a category. For example, we observe when training our models that the teapot spout that points away from the teapot body on the reconstructed object should likely align with the equivalent section of teapot spout on the observed object point cloud. We therefore extend our inference optimization procedure with an additional descriptor generated at training time, indicating relationships between object parts. For pairs of part point clouds on the same canonical object, we calculate the nearest-neighbor distance between points for each $x$ in point cloud $X$, $Dist(m,n) =  x^{(m)} - KNN(x^{(m)}, X^{(n)}, 1).$ We add a binary label to each point in both the canonical and observed part point clouds, $L^{(m)}_n = 1(dist_x < median(dist(m,n)) \times r )$, where $r=.4$ is an empirically-selected coefficient determining what portion of the point cloud receives each descriptor. We add an additional label  $L^{(m)}_z = 1(x_{z,i} < mean(X_z))$ associating points along the world's $z$-axis in their initial pose for further symmetry breaking. These labels are trivially identified on the observed demonstration and novel point clouds at test time. Our updated objective is then the sum of the Chamfer distance between only points sharing labels. 
\begin{align}
    \mathcal{D}\left( X, Y, L_m \right) = \sum_{i}^{(0,1)}{
         \frac{ \sum_{x}^{|\pcx{L_m = i}|}{\min \norm{ x - Y^{(L_m = i)}}^2}}{|L_m = i|}.
     }
    \label{eqn:chamfer}
\end{align}
We provide details on how multiple part relationships are handled simultaneously in \ref{multiple_descriptors}.

\begin{figure}
\captionsetup{width=\columnwidth}
\includegraphics[width=.9\columnwidth]{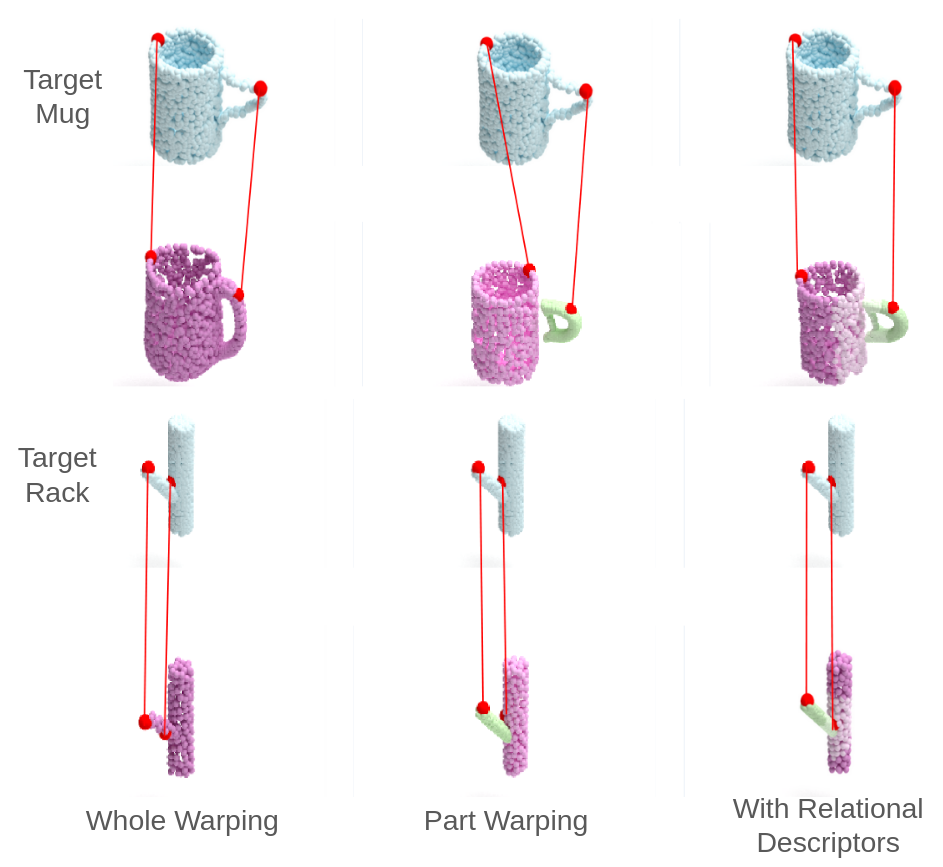}
\caption{The red points represent example keypoints $I^{(A)}$ on an object, the union of all keypoints $I^{(M)}$ of that object's parts. Whole object shape warping fails to adequately match the geometry of the object or the positions of the keypoints. Parts-based shape warping (PSW) without relational descriptors fails to correctly align the mug handle and to correctly position the annotated keypoints due to the rotational symmetry of the cup and trunk. PSW with relational descriptors is better able to reconstruct both geometry and keypoints.  }
\label{reconstruction_improvement}
\vspace{-3ex}
\end{figure} 

\subsection{Identifying salient part relationships}

Once $I^{m'}$ and $I^{n'}$ have been identified, given the neighbor indices and displacements extracted from the provided demonstration, we can find the transformation for each pair that best recreates the demonstrated alignment of object parts. For our experiments, we use Iterative Closest Point matching \cite{icp} to identify a $T_{m',n'}$ that best aligns interaction points. Notably, this results in a correct alignment of object \textit{parts}, but does not uniquely define a transformation between \textit{objects}. Naive transfer does not account for how the relative poses of those parts may have changed---for example, if the handle on the novel teapot were placed higher than the handle on the demonstration teapot. This is a common source of failure in whole object alignment, and may result in contradictions when considering keypoints representing all object part relationships simultaneously---a higher peg on a mug rack will change the relative reconstructed transformation between the handle and peg, but not between the handle and trunk, requiring further disambiguation.
Our approach addresses this disambiguation in two stages: first, constructing a heuristic as described in Equation \ref{heuristic_parts} by identifying which part relationships are relevant to successful reconstruction of the skill demonstration. And second, optimizing over that heuristic to obtain a transformation of the whole object's observed point cloud that aligns it with the point clouds of the relevant transformed parts. 
We will describe the second step, as we use it to validate the first. Given a set of relevant part relationships, $(m,n)_r$, we transform each partial point cloud $X^{(m')}$ by $T_{m'n'_r}$. We then optimize for a whole point cloud transformation, $T_{A'B'}$ that minimizes the heuristic objective described in equation \ref{eqn:chamfer} between each part $X^{(m')}$ present in a relevant part relationship, and the transformed reconstruction $Y^{(m')}$. If, for example, the body and spout of a teapot's relationship to the cup of a mug are both relevant, then the optimization objective is to minimize:
$$\sum^{part}_{(spout, body)}D(T_{teapot, mug}X^{(part)}, Y^{(part)})$$
while the relative pose of other parts (e.g. lid), are ignored.

To select the relevant part relationships for optimization, given our demonstration point clouds $X^{(M)}$ and $X^{(N)}$ and demonstration transformation $T_{AB}$, we generate all possible combinations of parts relationships. We run our inference procedure on the demonstration trajectory given each possible set of relevant parts, and score that set by how accurately it recreates $T_{AB}$ given the above optimization objective. We use that set of relevant relationships for all future instances of skill transfer. We consider small $M$ and $N$ in this work, so evaluating all part combinations is feasible. Tree pruning or program induction techniques could be useful for making larger search spaces manageable.

\section{Experiments} \label{segmethods}
We evaluate our method in two ways. First, we compare to existing benchmarks for manipulation transfer. Second, we compare performance across a range of different objects with different geometries. We evaluate on two object rearrangement tasks in simulation, and demonstrate performance on an additional task on the real robot, tea pouring, using an object (teapot) \textit{not present} in the Shapenet dataset \cite{shapenet}, to demonstrate the usefulness of sample efficiency in training. We also describe the perception system utilized for acquiring parts-segmented point clouds for real robot experiments. 

\textbf{Simulation:}
In open-source simulated environment Pybullet \cite{coumans2021}, we compare benchmark performance to our method on two tasks including at least one object with multiple constituent parts: placing a mug on a rack, and placing a bowl on a mug. Success is determined by fulfilling the task objective without clipping, measured through the normal force threshold on contacts between the object meshes immediately after placement. We provide a single demonstration and evaluate on randomly selected objects from a fixed test set in random initial starting poses, and compare the performance of our method against whole object interaction warping and relational neural descriptor fields \cite{simeonov21ndf}. We exclude local neural descriptor fields \cite{chun2023local} from this comparison as that work does not compare over shape variation in both objects involved in the skill. Results are in Table \ref{sim_results_table}---parts-based shape warping improves performance for both tasks.

\begin{figure*}[h]
\centering
\includegraphics[width=.90\textwidth]{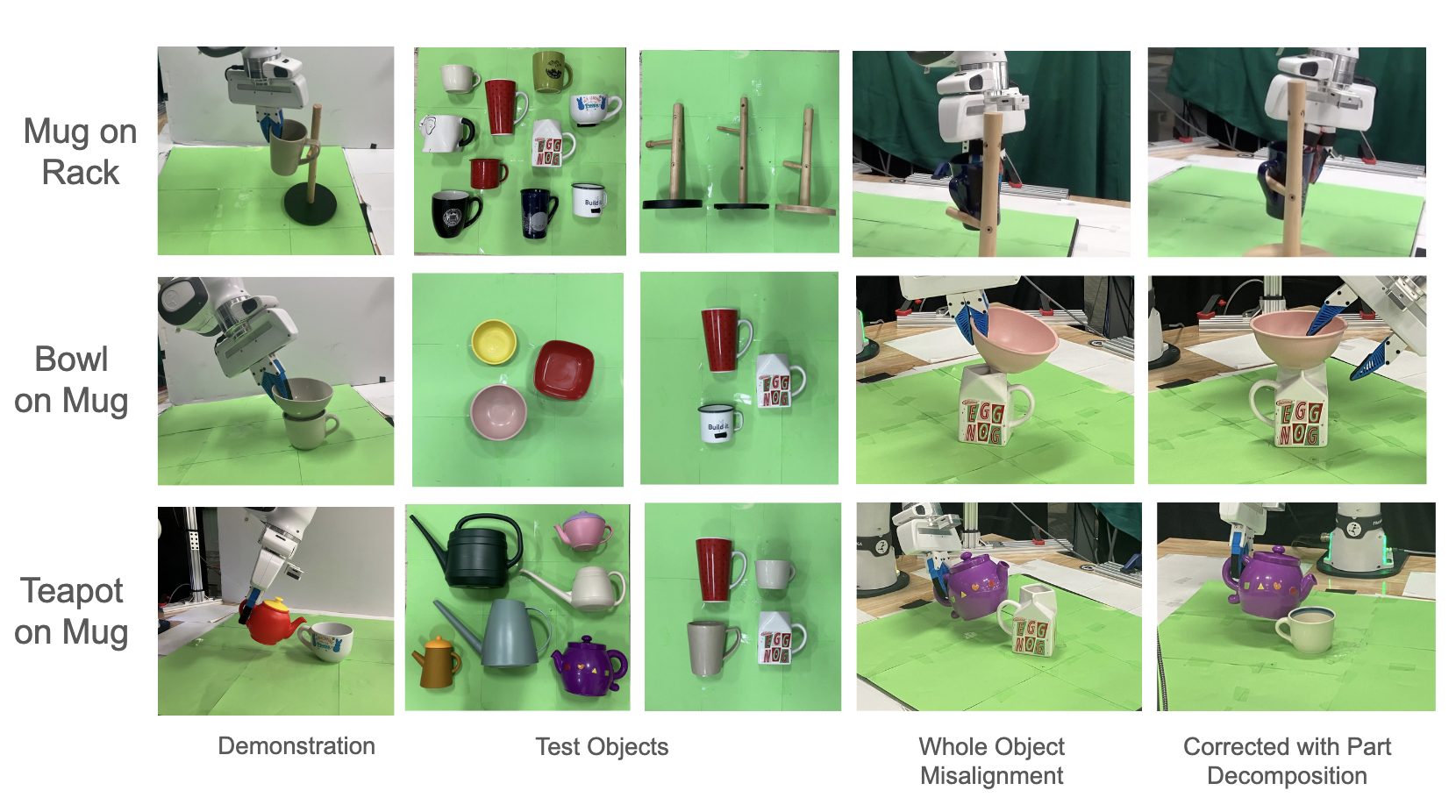}
\caption{We evaluated this method on objects of diverse geometry. The most common failure modes for whole object warping (IW)\cite{biza23oneshot} were misalignments due to poor geometric reconstruction (estimating the mug hook's position too high, for example) or skew from irrelevant geometric features (placing the teapots or bowls relative to a mug's handle, rather than its rim).}
\vspace{-3ex}
\end{figure*}

For the second evaluation, we focus on the task of placing a mug on a mug rack. We fix the rack to a particularly difficult example, with a narrow distance between the branch and trunk of the rack, resulting in frequent failure due to mesh intersection. Providing the same demonstration to each method, and providing an initial state in a random pose, we evaluate performance when placing every mug not used during training for each method. We visualize the results in Figure \ref{fig:opener}, which shows that our method succeeds on a wider range of objects of diverse geometries than whole object warping \cite{biza23oneshot} or R-NDF \cite{simeonov21ndf} even when only provided with 5 objects as training examples and a single demonstration. Compared to whole object warping, parts-based shape warping results in more accurate transfer of interaction points and better pose alignments, reducing the number of failures due to mesh collisions.

\begin{table}[h]
\begin{center}
\begin{tabular}[t]{p{.2\columnwidth}p{.18\columnwidth}p{.18\columnwidth}p{.18\columnwidth}}\toprule
 & RNDF\cite{simeonov21ndf} & IW\cite{biza23oneshot} & PSW \\
 \midrule
 Mug on Rack & $0.42 \pm 0.03$ & $0.64 \pm 0.03$ & \bm{$0.78 \pm 0.02$} \\
 Bowl on Mug & $0.43 \pm 0.03$  & $0.79 \pm 0.02$ & \bm{$0.84 \pm 0.02$}\\
 \bottomrule
\end{tabular}
\captionof{table}{Success rate (and standard error) of predicted placements in simulation, given one demonstration and five example object meshes, over 300 total evaluations.}\label{sim_results_table}
\end{center}
\vspace{-4ex}
\end{table}

\textbf{Real Robot:} We evaluate three tasks on the real robot: placing a mug on rack, placing a bowl on mug, and a pre-pouring alignment from a teapot into a mug. For each task, we provide a single demonstration in the form of an end-effector transformation and observations from four RGB-D cameras. We segment the object parts using the system described below and identify interaction points for the skill. Given RGB-D observations of a novel scene, we then warp the interaction points to identify a pre-place and final placement pose, using the same generative shape models as in simulation. The robot executes the skill by motion planning to those poses in sequence.  

We compare IW and PSW. Evaluating both methods requires segmenting object and part point clouds from real-world observations, handling noise and occlusions. 
For our experiments, we arbitrarily select a camera view of the skill demonstration where both grasped and target objects are visible. Along with the demonstration trajectory, the demonstrator provides 2-5 segmentation point prompts on one image for each part or whole object, and extracts a masked image using an existing segmentation model \cite{kirillov2023segment}. To acquire masks on the other camera views, we compare the cosine similarity of semantic features \cite{dinov2learningrobustvisual} across the selected image and each remaining camera's image, selecting the top $k$ ranked patches as point prompts on the novel views.

At test time, we perform this same process on all camera views of the novel objects. We aimed to evaluate these methods independent of the failures of a specific segmentation pipeline. If the masked point cloud produced was insufficient---lacking one of the object parts, or with substantial overlap between part segments---we provide additional point prompts on a single additional camera view of the novel objects and repeat the above procedure. The number of trials that required this resegmentation for each task are described in Table \ref{real_results_table}. This method produces part segmentations of objects of widely varying geometry and appearance.

We also demonstrate how this method enables cross-category generalization for objects that share sets of constituent parts. For our final experiment, we evaluate a set of watering cans as teapots, which share the subset of task-relevant teapot parts. In our experiments demonstrating cross-category transfer, we provide cross-category part equivalencies, as identifying them autonomously is beyond the scope of this evaluation, and will be explored in future work.

\begin{table}[h]
\begin{center}
\begin{tabular}[t]{lp{.12\columnwidth}p{.12\columnwidth}p{.12\columnwidth}p{.12\columnwidth}}\toprule
           & Parts (PSW) & Whole (IW) & Reseg (PSW) & Reseg (IW) \\\midrule
Mug/Rack    & 23/27 & 11/27 & 8/27 & 9/27  \\
Bowl/Mug & 8/9 & 6/9 & 2/9 & 1/9 \\
Teapot/Mug & 6/9 & 2/9 & 6/9 & 6/9 \\
\hline
Watering Can/Mug & 6/9 & - & 9/9 & - \\\bottomrule
\end{tabular}
\captionof{table}{Success rate of real-robot evaluations. Each evaluation is a unique pair of objects. }\label{real_results_table}
\end{center}
\vspace{-3ex}
\end{table}

We find that our method improves generalization of manipulation skills to objects of novel geometry, as shown in Table \ref{real_results_table}. Part decomposition allows the robot to better model objects with substantially different geometry from both the demonstration object and training meshes, and to condition on only relevant features for the given task, avoiding errors that contribute to object misalignment or collision. We note that the majority of failures of our method in the mug-on-rack task occurred with mugs with thin handles relative to their cups, leading to sparse point cloud observations. This biased the output of the final pose optimization towards denser areas of the point cloud due to the use of chamfer distance as the objective. A more refined metric may be able to avoid these failures. Other failures were largely attributable to imperfect reconstruction of object geometry resulting in collision or misaligned interaction points---particularly in the case of the narrow teapot/watering can spouts.

\section{Related Work}

\textbf{Skill transfer across objects of different shapes:} Existing work on skill transfer to novel objects utilize a range of different shape feature representations. 
\cite{manuelli19kpam} and visual flow techniques like \cite{zhu2024orion} utilize visual keypoints from models trained on significant amounts of data, transferring a skill by optimizing the alignment of keypoints identified on novel objects at test time. \cite{liu2024magic} identifies and localizes keypoints from contact points, and transfers them by matching semantic features and shape on a novel object. \cite{demo} and \cite{twobytwo} train object representations from ynthetic data. Our approach uses partial pointclouds, not images (though we use images for extracting segmented pointclouds). It identifies key interaction points in an unsupervised manner that does not require synthetic or labeled data or use of a pretrained model. 

R-NDF \cite{simeonov21ndf} foregoes explicit point-based feature representations in favor of a field-based object representation, predicting descriptors in 3D space to align for transfer. RelNDF works on whole objects, and requires on the order of hundreds of example object point clouds to train the neural descriptor field representation. It requires 5-10 demonstrations to describe a skill to be able to transfer to a new object. Our model requires 5-10 pose-aligned object meshes and 1 demonstration to achieve skill transfer. Rather than constructing our alignment descriptors from an existing model, we use symmetry-breaking heuristics to maintain alignment of object parts, and otherwise rely solely on geometry matching and the alignment of interaction points.

\textbf{Part-based compositional policies for manipulation skill transfer:} Composable Parts-Based Manipulation \cite{liu2024composablepartbasedmanipulation} models a manipulation skill as a set of constraints on object part relationships. These object-part relationships are trained from many engineered examples in simulation, and are predefined to exist between specific object parts. 
The final pose of the manipulated object is generated through sampling that optimizes to minimize the cost of these composed constraints. This work allows for cross-category transfer by identifying shared part relationships across demonstrations of different skills. Our work identifies part relationships from a single demonstration, and achieves transfer from a single demonstration, and does not focus on cross-category transfer. 

Local R-NDF \cite{chun2023local} and realtime-execution focuesed \cite{riemann} are extensions of  R-NDF where descriptors are associated with local geometry. This work doesn't assume access to semantic part categories or a segmentation method, which our work does. Like R-NDF, this work requires several demonstrations and many more objects to work, which our work does not.
\section{Conclusion and Limitations }
\label{sec:conclusion}
We show that parts-based shape warping improves transfer to objects of novel geometries with interaction warping. We demonstrate that this method enables one-shot transfer of multiple skills using a single demonstration and very few object examples on a real-world system.

\textbf{Limitations}: 
Identifying relevant object part relationships with a single demonstration sometimes results in mistaken association---a part relationship that seems relevant for that demonstration may not be relevant in all cases. This can be addressed by evaluating on more demonstrations, or using other metrics to identify relevant relationships. The shape-warping method used requires parts-segmented point clouds for each object involved in a task. Obtaining these from existing segmentation models proved sensitive to object appearance and environmental conditions, depending on the coverage of the model used. Segmentation errors may result in misidentification of features, leading to downstream planning failure. We'd expect performance to improve as more accurate models for segmentation become available. 

\section*{APPENDIX}

\subsection{Transferring Keypoints via Shape Warping} \label{iwstuff2}

Given interaction points  $I^{(m)}$ and $I^{(n)}$, and partial views of our novel object parts $X^{(m')}$ and $X^{(n')}$, a model is required to determine transformation $f$ to the equivalent positions of said annotations on the novel parts, $I^{(m')}$ and $I^{(n')}$. We achieve this by following a shape-warping method \cite{biza23oneshot}, which learns a generative model for inferring object geometry and the positions of interaction points on novel objects given a partial point cloud view. 

Rather than represent whole objects with a single model, we train a generative model per object part. We provide a pose-aligned set of 5-10 point cloud segments $\{\pcx{1}, \dots, \pcx{K}\}$ representing part instances belonging to class $M$. We select a 'canonical' part $\pcx{Canon}, Canon \in \{1,2, ..., K\}$ and define a set of displacement matrices 
$$\wxy{Canon}{j} = \mathrm{CPD}(\pcx{Canon}, \pcx{j}), j \in \{1, 2, ..., K\}$$ 
that represent a nonrigid registration between the canonical part and each $j$th part instance, using Coherent Point Drift \cite{myronenko10pointset} The choice of $Canon$ is arbitrary; we choose one most similar to all other object examples by Chamfer distance. 

We calculate a $d$-dimensional low rank approximation of the space of object-shape deformations using PCA. This approximates part shape using a low-dimensional latent vector $v_{\mathrm{novel}} \in \mathbb{R}^d$, which can be used to compute a transformation to the canonical point cloud representing a point cloud of a novel shape, $Y = \pcx{Canon} + \mathrm{Reshape}(W v_{\mathrm{novel}})$, where the $\mathrm{Reshape}$ operator casts back to an $n \times 3$ matrix.

There are two primary benefits to the above approach. First, it produces a latent space that captures meaningful shape variation across a part category from an accessible number of examples. Second, the low-rank latent space can be used as a generative model for approximating a part's complete shape from a partial point cloud view, naturally handling occlusion. We jointly infer the latent vector that best reconstructs the observed points as well as the pose of the object part in the world frame by minimizing an objective representing the difference between the target and reconstructed point clouds, as described in section $D$.

\begin{figure}
\label{relational_descriptors}
\captionsetup{width=\columnwidth}
\includegraphics[width=.88
\columnwidth]{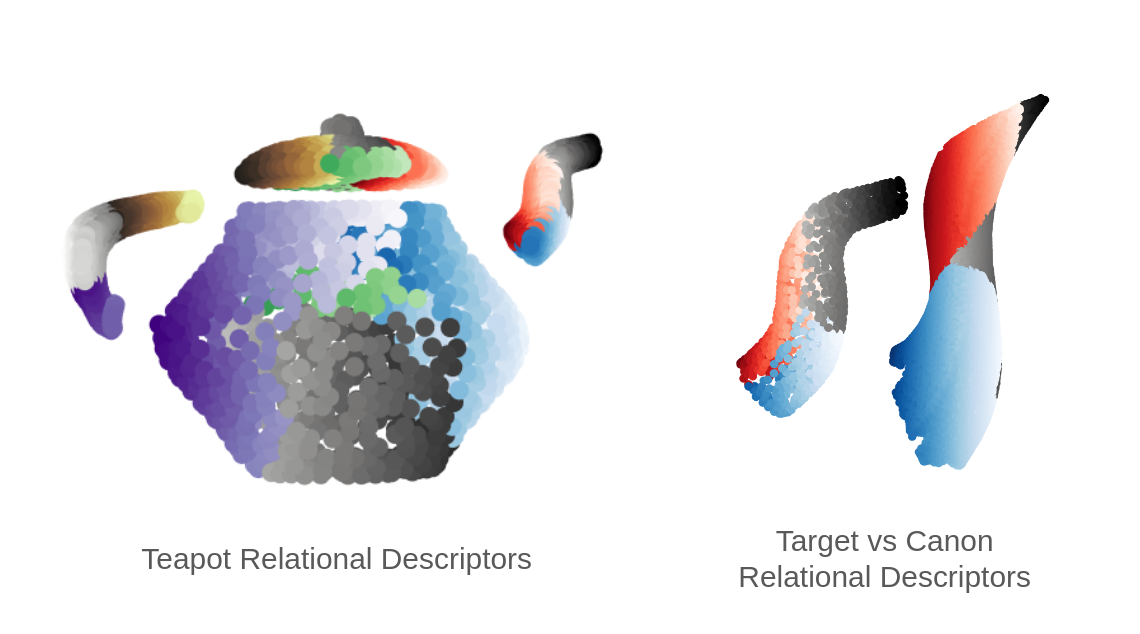}
\caption{An example of relational descriptors on a teapot. Part adjacency relationships are shown by matching colors annotated on the point clouds. The right shows an example of alignment between the canonical and target point clouds. }
\vspace{-4ex}
\end{figure}

\subsection{Relational Descriptors for Multiple Part Relationships}
\label{multiple_descriptors}

One part may have relevant adjacency relationships with multiple other parts---the body of a teapot is adjacent to the spout, lid, and handle. We determine adjacency between parts with a fixed threshold on the distance of closest points. For a set of adjacent parts $Adj(M,N)$, we define the objective function for shape reconstruction and alignment as the sum of objectives representing each part relationship, creating:
\begin{equation}
    \sum_{L_{(m,n)}}^{Adj(M,N)}{
             \mathcal{D}\left( \pcx{m}, Y^{(m)}, L_{(m,n)} \right)+ (\theta_{mean} - \theta_{X})
     },
    \label{eqn:sum_chamfer}
\end{equation}
where the final term is a regularization term that penalizes vectors in the latent space $\theta$ that are far from the mean of the latent components. This constrains the kinds of shapes we can generate, but disincentives sampling extreme or physically infeasible part shapes. We find including this part relationship descriptor breaks symmetry and improves reconstructions, as seen in Figure \ref{reconstruction_improvement}, as well as speeding up optimization by reducing total distance calculations. 

\addtolength{\textheight}{-6cm}   

\section*{Acknowledgments}
  This work was supported in part by ONR REPRISM MURI N00014-24-1-2603, ONR grant 00014-22-1-2592, the Robotics and AI Institute, and NSF GRFP 2439559.

\bibliographystyle{IEEEtran}

\bibliography{example}

\end{document}